\documentclass[10pt,twocolumn,letterpaper]{article}

\usepackage{iccv}              
\usepackage{times}
\usepackage{latexsym}
\usepackage{multicol}
\usepackage{natbib}
\usepackage{blindtext}
\usepackage{booktabs,chemformula,lingmacros,tree-dvips,multirow, notoccite}
\usepackage{epsfig}
\usepackage{graphicx}
\usepackage{amsmath}
\usepackage{amssymb}
\usepackage{array} 
\usepackage{hyperref}
\usepackage{placeins}

\usepackage{booktabs,chemformula,lingmacros,tree-dvips,multirow, notoccite}

\usepackage{microtype}

\newcommand{\rankone}{\textsuperscript{\textbf{\textcolor{blue}{1}}}}
\newcommand{\ranktwo}{\textsuperscript{\textbf{\textcolor{blue}{2}}}}
\newcommand{\rankthree}{\textsuperscript{\textbf{\textcolor{blue}{3}}}}

\NewDocumentEnvironment{alignb}{b}{%
  \begin{align*}
  \refstepcounter{equation} #1 \tag{\theequation}
  \end{align*}1
}{}
\def\paperID{8120} 
\def\confName{ICCV}
\def\confYear{2025}
\title{{S\textsuperscript{2}M\textsuperscript{2}}: Scalable Stereo Matching Model for Reliable Depth Estimation}

\author{
Junhong Min\textsuperscript{1, \dag}, Youngpil Jeon\textsuperscript{1}, Jimin Kim\textsuperscript{1}, Minyong Choi\textsuperscript{1}\\[0.25em] 
\textsuperscript{1}Samsung Electronics, \dag \textnormal{Corresponding author} \\[0.25em]
{ \{junhong1.min, yfeel.jeon, ji.min.kim, my312choi\}@samsung.com}
}
\begin{document}
\twocolumn[{%
\renewcommand\twocolumn[1][]{#1}%
\maketitle
\begin{center}
    \centering
    \captionsetup{type=figure}
    \includegraphics[width=.95 \textwidth]{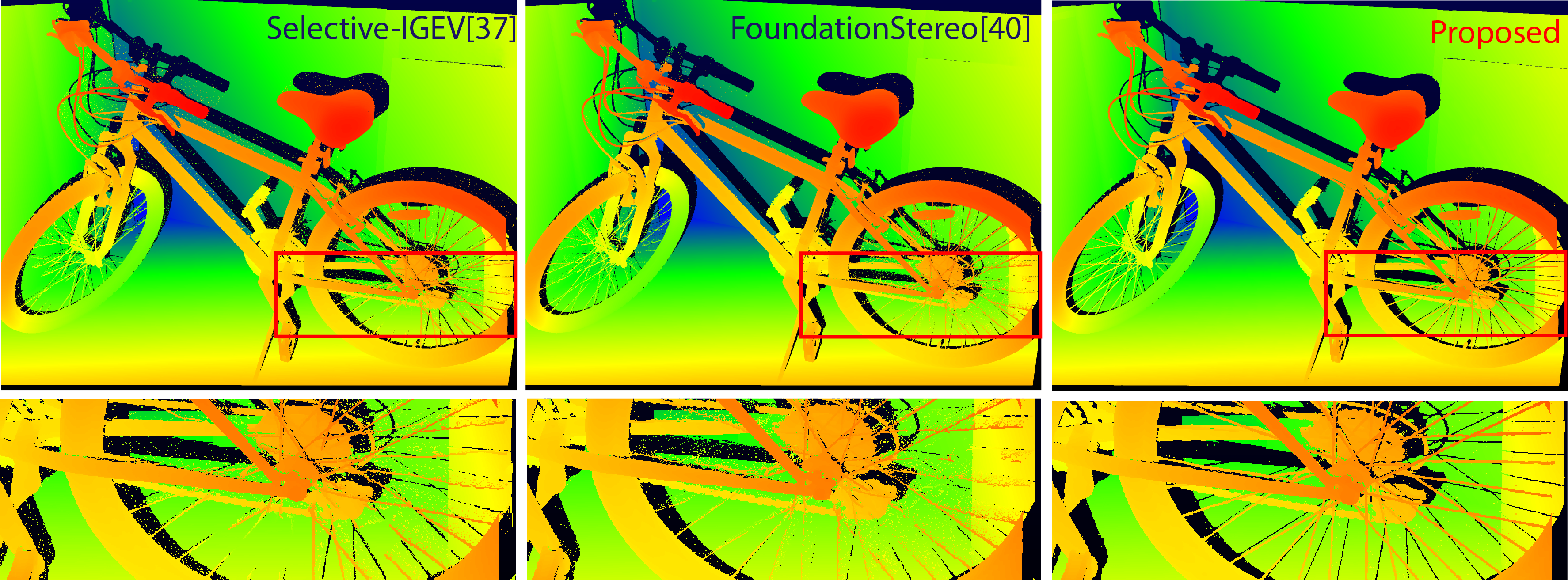}
    \caption{Qualitative comparison of 3D point clouds: Enhanced reliability in fine structures. 3D point cloud visualizations comparing our model against state-of-the-art methods on a challenging sample from the Middlebury v3 test set~\cite{scharstein2014high}. Our model yields more reliable reconstructions, especially in delicate structures like bicycle spokes, and reduces over-smoothing effects near edges.}
    \label{fig:thumbnail}
\end{center}%
}]

\begin{abstract}

The pursuit of a generalizable stereo matching model, capable of performing well across varying resolutions and disparity ranges without dataset-specific fine-tuning, has revealed a fundamental trade-off. Iterative local search methods achieve high scores on constrained benchmarks, but their core mechanism inherently limits the global consistency required for true generalization. However, global matching architectures, while theoretically more robust, have historically been rendered infeasible by prohibitive computational and memory costs. We resolve this dilemma with S²M²: a global matching architecture that achieves state-of-the-art accuracy and high efficiency without relying on cost volume filtering or deep refinement stacks. Our design integrates a multi-resolution transformer for robust long-range correspondence, trained with a novel loss function that concentrates probability on feasible matches. This approach enables a more robust joint estimation of disparity, occlusion, and confidence. S²M² establishes a new state of the art on Middlebury v3 and ETH3D benchmarks, significantly outperforming prior methods in most metrics while reconstructing high-quality details with competitive efficiency.

\end{abstract}    

\section{Introduction}
Recent advances in deep learning have significantly improved stereo matching accuracy, enabling high-quality depth estimation in well-constrained conditions \cite{chang2018pyramid, tankovich2021hitnet, badki2020bi3d, tosi2021smd, lipson2021raft, li2022practical, xu2023iterative, wang2024selective, min2024confidence}. However, ensuring reliable performance across diverse environments and image resolutions remains a challenge. A robust stereo matching model should generalize across different conditions (e.g., indoor/outdoor, low/high texture scenes) without requiring dataset-specific tuning. This pursuit of generalization often involves increasing model capacity for improved accuracy, but such gains do not always scale proportionally with size, requiring carefully designed architectures and training strategies. Furthermore, larger models inherently bring higher computational costs, necessitating a balance between efficiency and strong generalization, especially at high resolutions.

The landscape of deep learning-based stereo matching is primarily defined by a trade-off between two dominant architectural paradigms: global matching and iterative refinement. Transformer-based global matching architectures \cite{li2021revisiting, xu2023unifying, weinzaepfel2022croco, weinzaepfel2023croco, min2024confidence} are particularly well-suited for modeling long-range correspondences to handle occlusions and repetitive patterns. However, they have historically been underexplored because they pose significant efficiency challenges and are difficult to scale, both in processing high-resolution images and in effectively increasing model capacity. This scalability bottleneck led to the dominance of cost volume based iterative refinement methods \cite{lipson2021raft, zhao2022eai, li2022practical, xu2023iterative}, which demonstrated strong performance on established benchmarks. This paradigm, however, faces two fundamental limitations. First, processing the initial cost volume requires computationally expensive filtering, such as 3D convolutions, which severely hampers scalability. Second, the subsequent iterative refinement, being local in nature, struggles to resolve complex and discontinuous disparity structures, often failing to escape local minima.

To overcome the limitations inherent in both dominant paradigms, we propose the Scalable Stereo Matching Model (S²M²), which revitalizes the global matching approach by directly tackling its well-known scalability issues. S²M² introduces a multi-resolution transformer architecture that enables robust feature interactions both across different scales and within each resolution level, ensuring stable learning as the model scales efficiently. Additionally, an adaptive feature fusion mechanism selectively integrates multi-resolution information, preserving fine details while maintaining global consistency. These combined enhancements allow S²M² to achieve robust depth estimation across varying image resolutions and scene complexities.

In addition to the architecture, we introduce a dedicated loss function designed to guide probability concentration on feasible matches. By explicitly encouraging the network to focus probability mass on valid candidates, our model not only achieves more precise disparity predictions but also jointly produces a corresponding confidence score derived directly from this distribution. This allows for robust filtering of uncertain predictions in ambiguous areas, such as occlusions and textureless surfaces \cite{poggi2021confidence, costanzino2023learning}, which is crucial for reliable performance in safety-critical applications \cite{eitel2015multimodal, murali2020learning, kendall2017end}.

\begin{figure}[!htbp]
    \centering
    \includegraphics[width=1.0\linewidth]{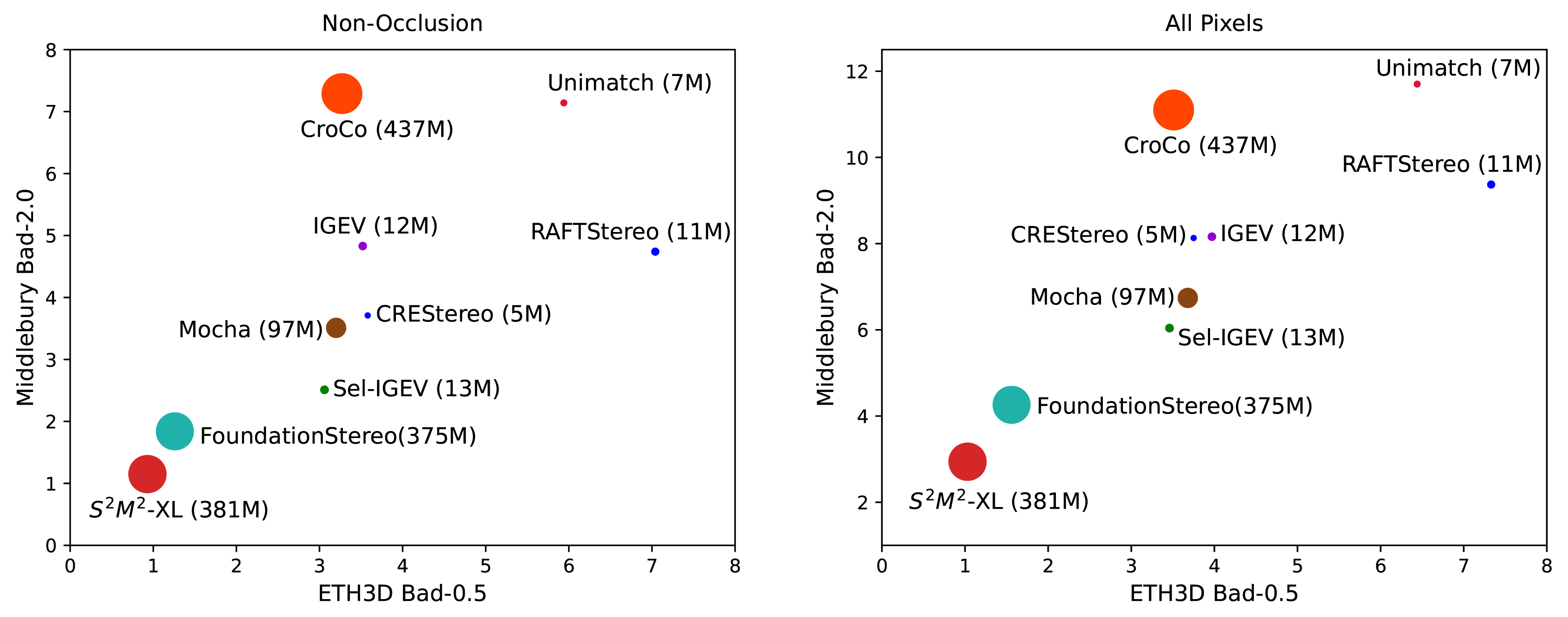}
    \vspace*{-1mm}
    \caption{Comprehensive evaluation on two representative benchmarks, ETH3D~\cite{schops2017multi} and Middlebury v3~\cite{scharstein2014high}. Accuracy is measured using the Bad-2.0 metric for Middlebury and Bad-0.5 for ETH3D. Each method is plotted with its corresponding model size (in millions of parameters).}
    \label{fig:benchmark_summary}
\end{figure}

The performance of our model was validated on challenging real-world benchmarks, conducting comprehensive comparisons against state-of-the-art methods. As Figures~\ref{fig:thumbnail} and \ref{fig:benchmark_summary} illustrate, our approach demonstrates strong visual and quantitative performance across diverse real-world datasets~\cite{scharstein2014high, schops2017multi}. While real-world benchmarks are indispensable for evaluating stereo matching, they inherently present limitations such as restricted disparity ranges or imperfect ground-truth annotations, especially in complex or fine-structured regions. To complement this real-world evaluation, we additionally constructed a high-fidelity synthetic dataset, which enables a more thorough assessment under controlled conditions, including scenes with larger disparity variations and precisely defined ground truth. By achieving state-of-the-art results on both real-world and synthetic benchmarks, our work is the first to conclusively demonstrate that a carefully designed global matching architecture can be scalable in terms of both input resolution and model capacity, overcoming its well-known limitations to achieve robust and state-of-the-art performance.

To summarize, our key contributions include:
\begin{itemize}
    \item \textbf{A Highly Scalable Global Matching Architecture:} A multi-resolution Transformer architecture scalable in terms of both input (high resolutions, large disparities) and model size.

    \item \textbf{Accurate and Reliable Depth Estimation:} A novel loss function that boosts disparity accuracy, while the joint estimation of confidence and occlusion ensures final depth reliability.

    \item \textbf{New SOTA and a Rigorous Validation Framework:} State-of-the-art performance on established real-world benchmarks, and a novel synthetic dataset to validate robustness in scenarios unaddressed by existing benchmarks.
\end{itemize}

\section{Related Works}
\begin{figure*}[!htbp]
    \centering
    \includegraphics[width=.95\linewidth]{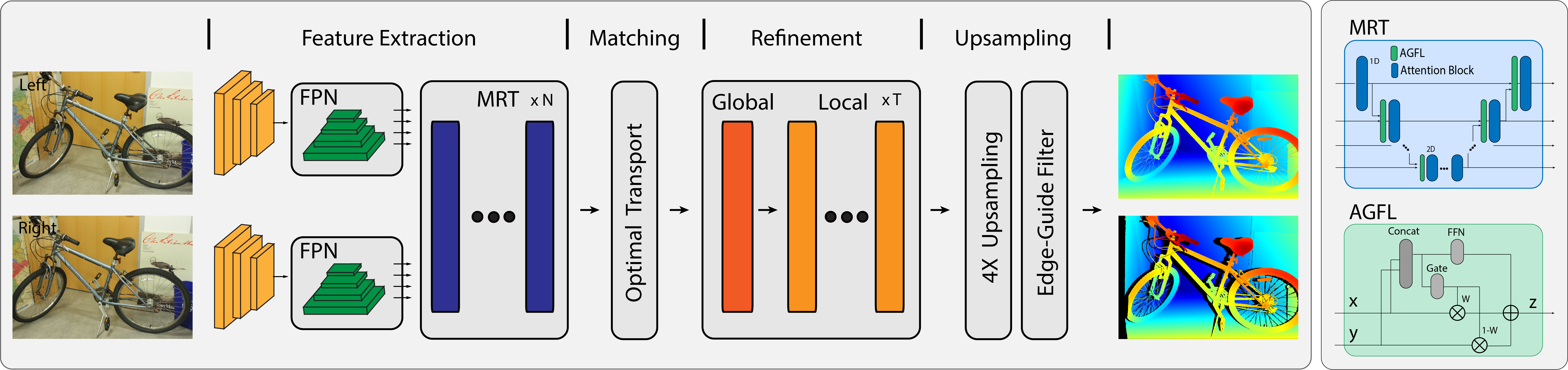}
    \caption{Our proposed architecture, S2M², comprises four main stages: (1) feature extraction, where left and right input images are processed by FPN and MRT to yield multi-resolution feature maps; (2) global matching, which computes all-pairs correlation at 1/4 resolution and estimates disparity, occlusion, and confidence using optimal transport; (3) refinement, iteratively updating these estimates; and (4) upsampling, enhancing depth boundaries to restore full resolution. The feature extraction stage employs a Multi-Resolution Transformer (MRT) for hierarchical feature extraction, incorporating an Adaptive Gated Fusion Layer (AGFL) for efficient multi-scale feature alignment.}

    \label{fig:network-overview}
\end{figure*}

The era of deep learning in stereo matching began with CNN-based methods~\cite{kendall2017end, duggal2019deeppruner}, which effectively replaced handcrafted components of classical algorithms like SGM~\cite{hirschmuller2008sgm} with powerful deep feature extractors. While these models significantly improved accuracy by learning robust feature representations, their reliance on local convolutions resulted in limited generalization across diverse domains and resolutions. This limitation became a critical bottleneck, necessitating the exploration of new architectural paradigms.

Modern stereo matching has largely diverged into two parallel pathways: Transformer-based global matching and cost volume-based iterative refinement. The former, Transformer-based methods, leverage self- and cross-attention to establish robust global correspondences, making them particularly effective for handling occlusions and repetitive patterns. An early model, STTR~\cite{li2021revisiting}, provided foundational insights by framing the task as a sequence-to-sequence problem, though with limited scalability. More recent approaches~\cite{weinzaepfel2022croco, weinzaepfel2023croco, xu2023unifying, min2024confidence} have refined their global correspondence mechanisms by incorporating techniques from related fields such as dense keypoint matching~\cite{sarlin2020superglue,sun2021loftr,wang2022matchformer} and point cloud registration~\cite{yew2022regtr,yu2023rotation,qin2022geometric}. Nonetheless, the quadratic complexity of attention mechanisms remains a critical bottleneck, hindering scalability in terms of both input resolution and model capacity. This presents a key research gap that our work aims to address.

In parallel, cost volume based iterative refinement has emerged as a dominant pathway, demonstrating strong practical performance. This approach was pioneered by RAFT-Stereo~\cite{lipson2021raft}, which introduced recurrent updates on a learned correlation volume. Subsequent works have advanced this framework through several key strategies. For instance, CREStereo~\cite{li2022practical} introduced a more robust, hierarchical recurrent refinement to better handle fine details and real-world image distortions. Another line of work, such as IGEV~\cite{xu2023iterative}, incorporated geometric guidance to improve disparity propagation. More recently, efforts have centered on refining feature representations, with Selective-IGEV~\cite{wang2024selective} selectively integrating frequency information and MoCha-Stereo~\cite{chen2024mocha, chen2024motifv2} employing motif-based attention for robustness. Despite these advancements, the performance of these methods hinges on a high-quality correlation volume produced by computationally expensive filtering, and the inherently local nature of this process makes them less effective in globally ambiguous regions with repetitive textures or large discontinuities, reinforcing the need for true global correspondence mechanisms.

Another distinct line of research has focused on computationally efficient methods for real-time applications~\cite{bangunharcana2021correlate, xu2021bilateral, xu2020aanet, xu2023accurate}. These approaches typically achieve efficiency by replacing expensive 3D cost aggregation with lightweight, 2D-based correlation or spatial propagation networks. While they achieve competitive accuracy in low-resolution settings suitable for real-time use, their design often leads to a significant performance degradation on high-resolution imagery. This reveals a fundamental trade-off between efficiency and high-resolution accuracy, a gap that our scalable architecture is designed to bridge.

A recent trend aims to improve generalization by leveraging large-size, pre-trained models. These methods typically build upon the iterative refinement framework but replace the standard feature encoder with a powerful backbone pre-trained on other large-scale tasks. This is often achieved by incorporating a general purpose vision encoder or even a complete monocular depth estimation model as a feature extractor~\cite{zhou2025all, wen2025foundationstereo, jiang2025defom, bartolomei2025stereo}. By transferring rich prior knowledge, these approaches demonstrate impressive generalization capabilities. However, this strategy introduces a significant trade-off: their reliance on massive, often fixed-weight components results in models that are too large and computationally expensive for many practical applications. This highlights the ongoing challenge of achieving both strong generalization and efficiency, a balance our scalable architecture aims to achieve.

\section{Method}

Our proposed model, S²M², is designed to revitalize the global matching paradigm by addressing its long-standing scalability challenges. To achieve this, our architecture is composed of four main stages, as illustrated in Figure~\ref{fig:network-overview}: (1) a feature extractor that leverages a multi-resolution architecture to produce a single, powerful feature map; (2) a global matching module that uses optimal transport~\cite{cuturi2013sinkhorn} to establish robust long-range correspondences; (3) a hybrid refinement stage that corrects errors both globally and locally; and (4) a final upsampling process to restore full resolution while preserving details.

\subsection{Feature Extraction}
The goal of our feature extraction stage is to produce a single, high-resolution feature map for each image, rich with the contextual information necessary for robust global matching. While using a large-scale Vision Transformer for this task is a common approach, it is computationally prohibitive for high-resolution stereo, creating the primary scalability bottleneck. To address this, we propose an efficient hierarchical architecture. The process begins with a lightweight CNN backbone and a Feature Pyramid Network (FPN) that generate an initial multi-scale feature pyramid: $\{\mathbf{F}_4, \mathbf{F}_8, \mathbf{F}_{16}, \mathbf{F}_{32}\}$. These features are then fed into our novel Multi-Resolution Transformer (MRT) for deep contextual refinement.

\subsubsection{Multi-Resolution Transformer (MRT)}
The MRT is designed to overcome the efficiency limitations of conventional Transformers. Instead of processing a single high-resolution feature map, it operates on the entire feature pyramid in parallel. Its efficiency stems from a hybrid attention strategy: to align with epipolar geometry, attention is restricted to the horizontal axis (1D) at higher resolutions, while full 2D self-attention is applied only at the coarsest level ($\mathbf{F}_{32}$) to capture global context. This design strikes a critical balance between performance and computational cost. 

The internal structure of each MRT block, illustrated in Figure~\ref{fig:network-overview}, is also carefully designed. Each block is composed of a self-attention module and a cross-attention module, with each respective attention layer being followed by an FFN to produce highly discriminative features. To ensure stable training of a deep stack of MRT blocks, each block is also equipped with cross-scale skip connections.

\subsubsection{Adaptive Gated Fusion Layer (AGFL)}
We build a high-capacity feature extractor by stacking multiple MRT blocks. A key challenge in such a deep, multi-resolution architecture is ensuring effective information exchange and coherent learning across all scales. Prior architectures like U-Net can suffer from gradient bottlenecks at the highest resolution, hindering the training of coarser-level features~\cite{min2024confidence}. While our parallel MRT architecture avoids this specific bottleneck, it still requires a mechanism for robust cross-scale communication.

To facilitate this, we introduce the Adaptive Gated Fusion Layer (AGFL). The AGFL acts as a dynamic gate within each MRT block, learning to selectively fuse information between scales. This allows global context from coarser levels to effectively inform finer ones and vice versa, ensuring stable training and a powerful, coherent multi-scale representation. The fusion process is formulated as:
\begin{equation} 
    \mathbf{z} = \mathbf{w} \odot \mathbf{x} + (1 - \mathbf{w}) \odot \mathbf{y} + \text{MLP}_r(\text{Concat}(\mathbf{x}, \mathbf{y})) 
\end{equation} 
with the weight $\mathbf{w}$ computed as: 
\begin{equation} 
    \mathbf{w} = \sigma(\text{MLP}_w(\text{Concat}(\mathbf{x}, \mathbf{y}))).
\end{equation}

\subsection{Global Matching}
Given the enhanced feature maps from our feature extractor, the next stage establishes dense global correspondences. A common approach in global matching involves first constructing an all-pairs correlation volume $\mathbf{C} \in \mathbb{R}^{H' \times W' \times W'}$ from the 1/4-resolution features ($\mathbf{F}_4$), where $H' = H/4$ and $W' = W/4$. This volume encodes feature similarities along the epipolar lines. Following this standard practice in global matching, we formulate the task as a global assignment problem solved via entropy-regularized optimal transport~\cite{cuturi2013sinkhorn}, instead of relying on a simple \texttt{argmax} operation on this volume. By enforcing a soft, bidirectionally consistent (one-to-one) matching constraint, the Sinkhorn algorithm finds a globally optimal transport plan $\mathbf{T}$ that is robust to ambiguities where a simple best match from one perspective might not be the best from the other.
\begin{equation}
    \mathbf{T} = \operatorname{OT}(\mathbf{C}).
    \label{eq:ot}
\end{equation}
To handle occlusions, we incorporate the standard dustbin trick~\cite{sarlin2020superglue, min2024confidence}.

A key advantage of this formulation is that the resulting transport plan $\mathbf{T}$ provides a rich, probabilistic representation from which all initial estimates can be efficiently derived. The disparity map $\mathbf{D}$ is computed as the expected value over all candidates:
\begin{equation}
    \label{eq:global_disp}
    \mathbf{D}_{ij} = \frac{\sum_{w=1}^{W'} \mathbf{T}_{ijw} \mathbf{p}_w}{\sum_{w=1}^{W'} \mathbf{T}_{ijw}},
\end{equation}
where $\mathbf{p}_w$ represents the candidate disparities. The occlusion map $\mathbf{O}$ is obtained by marginalizing probabilities across the disparity axis, and the confidence map $\mathbf{\Gamma}$ is derived by summing the probability mass within a local window around the peak probability:
\begin{equation}
    \label{eq:global_conf}
    \mathbf{\Gamma}_{ij} = \sum_{w \in \mathcal{W}(w^*)} \mathbf{T}_{ijw},\quad\text{where}\quad w^* = \arg\max_{w} \mathbf{T}_{ijw}.
\end{equation}
These three estimates—disparity, occlusion, and confidence—serve as a robust starting point for the subsequent refinement stage.

\subsection{Refinement}

The refinement stage consists of two steps: a global disparity adjustment and an iterative local refinement process. The global step addresses occluded regions by propagating disparity values from high-confidence areas, following a confidence-aware disparity aggregation strategy~\cite{min2024confidence}. Given the initial disparity map $\mathbf{D}_0$ and confidence map $\mathbf{\Gamma}_0$, the global refinement step updates disparity in unreliable regions as  
\begin{equation}
    \mathbf{D}_0' = \mathbf{D}_0 + \Phi(\mathbf{F}_4^L, \mathbf{\Gamma}_0, \mathbf{D}_0),
\end{equation}
where $\Phi$ represents the aggregation function that primarily propagates disparity from high-confidence regions to low-confidence regions based on the confidence weights.

For local refinement, we iteratively update disparity, occlusion, and confidence maps using a residual correction strategy inspired by RAFT-style recurrent refinement~\cite{lipson2021raft}. At each iteration $t$, the updates are formulated as  
\begin{equation}
    \mathbf{D}_{t+1} = \mathbf{D}_t + \Delta \mathbf{D}_t,
\end{equation}
\begin{equation}
    \mathbf{O}_{t+1} = \mathbf{O}_t + \Delta \mathbf{O}_t,
\end{equation}
\begin{equation}
    \mathbf{\Gamma}_{t+1} = \mathbf{\Gamma}_t + \Delta \mathbf{\Gamma}_t.
\end{equation}
where $\Delta \mathbf{D}_t$, $\Delta \mathbf{O}_t$, and $\Delta \mathbf{\Gamma}_t$ are residual updates computed by the refinement network. These updates are conditioned on local feature correlations, disparity estimates, and the confidence map, ensuring that occlusions and unreliable disparities are iteratively corrected.

\subsection{Upsampling}
Following refinement, the low-resolution disparity map is upsampled to its original resolution. We first perform an initial $4\times$ upsampling using a weighted combination of neighboring disparity values, where the weights are learned from the feature maps, similar to prior works~\cite{tosi2021smd, li2022practical, xu2023iterative}. To further enhance the quality of object boundaries, we then apply a final edge-guided filter. This filter uses adaptive weights, predicted from the left image and a disparity-warped right image, to refine the upsampled disparity map. This final filtering step improves stereo consistency while preserving fine details at depth edges.

\section{Loss Function}
We train our model with a composite loss function including disparity, occlusion, confidence (all supervised by L1 loss with ground truths), and a Probabilistic Mode Concentration (PMC) loss. Standard L1 losses on disparity, confidence, and occlusion do not directly guide the Transformer's matching probabilities. To address this, we introduce the PMC loss, which encourages concentrated matching probabilities within valid disparity regions, enhancing feature discriminability and global matching robustness.

\subsection{Probabilistic Mode Concentration (PMC) Loss}
Global matching is performed on 1/4 downsampled feature maps, resulting in at most 16 disparity candidates at each pixel in this lower resolution space. Let $K$ (with $K \le 16$) denote the number of these candidates. For each candidate $c_{ij}^{(k)}$, we define its neighborhood as
\begin{equation}
    \label{eq:neighborhood}
    \mathcal{N}\bigl(c_{ij}^{(k)}\bigr) = \{ w \in \{1,\dots,W\} \mid |w - c_{ij}^{(k)}| \le \delta \},
\end{equation}
and take the union over all candidates:  
\begin{equation}
    \label{eq:union}
    \mathcal{U}_{ij} = \bigcup_{k=1}^{K} \mathcal{N}\bigl(c_{ij}^{(k)}\bigr).
\end{equation}  
The aggregated probability mass over this union is then computed as  
\begin{equation}
    \label{eq:prob_mass}
    S_{ij} = \sum_{w \in \mathcal{U}_{ij}} T_{ij}(w),
\end{equation}  
where $\mathbf{T}_{ij} = [T_{ij}(1), T_{ij}(2), \dots, T_{ij}(W')]$ represents the matching probability vector obtained from  Eq.~\ref{eq:ot}

The PMC loss enforces that $S_{ij}$ remains close to 1 for non-occluded pixels:  
\begin{equation}
\label{eq:loss_pmc}
    \mathcal{L}_{\mathrm{PMC}} = \frac{1}{|\Omega|} \sum_{(i,j)\in\Omega} \max(1 - S_{ij} - \epsilon, 0),
\end{equation}  
where $\Omega = \{(i,j) \mid O(i,j)=1\}$ represents the set of non-occluded pixels, and $\epsilon$ is a small margin.

This additional loss regularizes the cost volume, ensuring that the matching probability remains concentrated within valid candidate regions, thereby complementing the disparity loss and guiding the network toward confident and precise matches.

The final loss function combines disparity, confidence, occlusion, and PMC losses, each weighted by a hyperparameter:
\begin{equation}
    \label{eq:loss_total}
    \mathcal{L} = \lambda_D\,\mathcal{L}_D + \lambda_{\mathcal{O}}\,\mathcal{L}_{\mathcal{O}} + \lambda_{\mathcal{C}}\,\mathcal{L}_{\mathcal{C}} + \lambda_{\mathrm{PMC}}\,\mathcal{L}_{\mathrm{PMC}}.
\end{equation}  
\begin{figure}[!tbp]
    \centering
    \includegraphics[width=.75\linewidth]{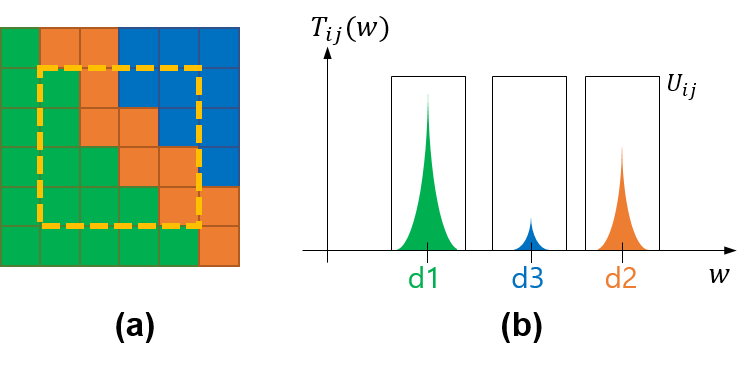}
    \caption{Illustration of Probabilistic Mode Concentration (PMC) Loss. (a) Disparity map with a dashed box highlighting a 4×4 region, which corresponds to a single pixel in the 1/4 downsampled feature space. (b) Matching probability distribution at the selected location, where PMC Loss ensures that the probability mass remains concentrated within valid disparity candidates.
}
    \label{fig:pmc}
\end{figure}

\begin{table*}[!htp]
\centering
\renewcommand{\tabcolsep}{3pt} 
\footnotesize{
\begin{tabular}{ccc|c|cc|cc|cc}
\hline\hline
\multicolumn{3}{c|}{\textbf{Model Configuration}} & \multicolumn{1}{c|}{\textbf{Training Strategy}} & \multicolumn{2}{c|}{\textbf{Disparity Metrics}} & \multicolumn{2}{c|}{\textbf{Auxiliary Metrics}} & \multicolumn{2}{c}{\textbf{Model Complexity}} \\
\hline
\textbf{Ch} & \textbf{Ntr} & \textbf{AGFL} & \textbf{PMC Loss} & \textbf{EPE} & \textbf{Bad-2.0} & \textbf{Occ AP} & \textbf{Conf AP} & \textbf{Params (M)} & \textbf{GFLOPs} \\
\hline\hline
128 & 1 & No  & No  & 0.489 & 1.974 & 0.973 & 0.977 & 23 & 645 \\
128 & 1 & Yes & No  & 0.496 & 1.601 & 0.973 & 0.978 & 24 & 647 \\
128 & 1 & Yes & Yes & 0.451 & 1.490 & 0.974 & 0.978 & 24 & 647 \\
128 & 3 & Yes & Yes & 0.388 & 1.281 & 0.974 & 0.978 & 43 & 772 \\
256 & 1 & Yes & Yes & 0.320 & 0.928 & 0.975 & 0.979 & 93 & 2146 \\
256 & 3 & Yes & No  & 0.299 & 0.810 & 0.976 & 0.980 & 170 & 2644 \\
256 & 3 & Yes & Yes & 0.254 & 0.750 & 0.976 & 0.980 & 170 & 2644 \\
\hline
\end{tabular}}
\caption{Ablation study on model configuration (ch, ntr, Adaptive Gated Fusion Layer (AGFL)) and training strategy (Probabilistic Mode Concentration (PMC) loss). Disparity metrics (EPE, bad-2.0) evaluate stereo matching accuracy, while auxiliary metrics (occ AP, conf AP) measure average precision (AP) for occlusion and confidence estimation. Model complexity is represented by params (M), the number of learnable parameters, and GFLOPs, which quantifies computational cost in giga floating-point operations per second (GFLOPs) for a $960\times512$ px input. ch: base channel dimension, ntr: number of transformer blocks.}
\label{tab:ablation}
\end{table*}

\begin{table}[h]
\centering
\footnotesize{
\begin{tabular}{c|c||cc|cc}
\hline\hline
\textbf{Iterations} & \textbf{GFLOPs} & \multicolumn{2}{c|}{\textbf{Disparity Metrics}} & \multicolumn{2}{c}{\textbf{Auxiliary Metrics}} \\
\cline{3-6}
& & \textbf{EPE} & \textbf{Bad-2.0} & \textbf{Occ AP} & \textbf{Conf AP} \\
\hline\hline
0 & 305  & 0.666 & 2.768  & 0.961 & 0.977 \\
1 & 427  & 0.470 & 1.658  & 0.974 & 0.977 \\
2 & 537  & 0.479 & 1.526  & 0.974 & 0.977 \\
3 & 647  & 0.451 & 1.490  & 0.974 & 0.978 \\
\hline
\end{tabular}}
\caption{Impact of local refinement iterations on computational cost and performance. Model Configuration (Ch=128,Ntr=1). Increasing the number of refinement iterations improves disparity estimation accuracy (EPE, Bad-2.0) and auxiliary metrics (Occ AP, Conf AP), but also increases GFLOPs.}
\label{tab:refinement}
\end{table}

\begin{table*}[!htp]
\centering
\footnotesize{
\resizebox{\textwidth}{!}{
\begin{tabular}{l|c c c c c c|c c c c c c |c c c}
\hline\hline
\multirow{2}{*}{\textbf{Model}} & \multicolumn{6}{c|}{\textbf{Non-Occlusion}} & \multicolumn{6}{c|}{\textbf{All Pixels}} & \multirow{2}{*}{\textbf{Mem(G)}} & \multirow{2}{*}{\textbf{TFLOPs}} & \multirow{2}{*}{\textbf{Param(M)}}\\ 
\cline{2-13} 
& \textbf{EPE} & \textbf{RMS} & \textbf{Bad-0.5} & \textbf{Bad-1.0} & \textbf{Bad-2.0} & \textbf{Bad-4.0} 
& \textbf{EPE} & \textbf{RMS} & \textbf{Bad-0.5} & \textbf{Bad-1.0} & \textbf{Bad-2.0} & \textbf{Bad-4.0} \\ 
\hline\hline
{CoEx} \cite{bangunharcana2021correlate} &20.17 &63.95 &38.78 &27.82 &21.06 &16.75 
&37.55	&103.82 &46.95 &36.78 &29.85 &24.87
&31.31 &0.71 &2.72
\\ \hline
{BGNet+} \cite{xu2021bilateral}&14.04	&42.84	&43.97	&29.55	&20.63	&14.82
&30.31	&88	&50.98	&37.85	&29.11	&22.87 
&22.49 &2.42 &5.31\\ \hline
{RAFT} \cite{lipson2021raft}&20.41	&48.12	&24.41	&17.71	&13.87	&11.32
&33.07	&76.51	&33.31	&26.43	&21.84	&18.31
&8.18 &30.35 &11.11\\ \hline
{fastACV} \cite{xu2023accurate} &15.09	&48.39	&36.73	&24.77	&17.69	&13.31
&32.50	&93.99 &45.57	&34.61	&27.34	&22.10
&8.01 &0.60 &3.07\\ \hline
{UniMatch \cite{xu2023unifying}} &6.46 &17.54 &55.77	&31.11	&16.35	&10.66
&13.96	&37.93	&61.77	&40.05	&25.82	&18.54
&49.3 &16.35 &7.35\\ \hline
{CroCo} \cite{weinzaepfel2023croco}&15.64	&35.66	&33.82	&23.71	&17.43	&13.48
&24.99	&57.23	&40.70	&30.96	&24.31	&19.57
&2.62 &164.7 &437\\ \hline
{IGEV} \cite{xu2023iterative}&7.03 &21.16 &23.04 &13.86 &9.66 &7.13	
&15.42	&42.85	&31.11	&22.17	&17.30	&13.82 
&10.87 &27.63 &12.59\\ \hline
{MC} \cite{feng2024mc} &35.56 &73.77 &32.48 &26.72 &23.08 &20.56
&56.07	&112.67 &41.12 &35.47 &31.46 &28.29
&7.73 &899 &24.62\\ \hline
{NMRF} \cite{guan2024neural} &40.88 &82.83 &58.67 &43.24 &36.66 &33.07
&62.04	&121.41 &64.06 &50.57 &44.28 &40.31
&9.07 &3.28 &6.113\\ \hline
{Sel-IGEV} \cite{wang2024selective}&9.23 &25.92 &22.37 &14.09 &9.79 &7.36
&18.23	&48.79	&30.06 &21.82 &16.86 &13.49
&13.4 &35.77 &13.14\\ \hline
{MoCha} \cite{chen2024mocha} &40.37 &102.65 &26.9 &21.62 &18.17 &15.73
&59.75	&141.19 &35.34	&29.91	&25.94	&22.82
&12.87 &22.14 &97.14\\ \hline

\textbf{S²M²-S} (+1Mtr) & 6.45 & 16.36 & 19.10 & 11.46 & 7.63 & 5.39
&11.39 &31.06 &26.94 &18.93 &  14.15 &10.78
&8.17 &5.76	&23.91\\ \hline

\textbf{S²M²-L} (+1Mtr)  &3.05\rankthree  &10.17\rankthree  & 14.92  & 8.72  & 5.58  & 3.74
 & 8.21\rankthree  & 25.92  & 23.48  & 16.53	 &  12.21  & 9.12
&11.39 &25.96 &170\\ \hline\hline

{FoundationStereo \cite{wen2025foundationstereo} (+2Mtr)}  &{4.60} &11.27  &{10.96}\rankthree & {7.23}\rankthree &{4.95}\rankthree &{3.49}\rankthree &
{8.26} &{23.04}\rankthree &{18.42}\rankthree &{13.98}\rankthree &{10.73}\rankthree &{8.08}\rankthree
&53.78 &124.65 &375\\ \hline

\textbf{S²M²-S} (+2Mtr) &4.82 &{10.83} &13.13 &7.62 &5 &3.57
&9.14 &26.32 &20.75 &14.55 &10.83 &8.26
&8.17 &5.76	&23.91\\ \hline

\textbf{S²M²-L} (+2Mtr)  &2.99\ranktwo  &8.72\ranktwo  &{10.60}\ranktwo  &{6.00}\ranktwo  &{3.80}\ranktwo  &{2.54}\ranktwo
 &{7.20}\ranktwo  &{22.40}\ranktwo  &{18.11}\ranktwo  &{12.72}\ranktwo	 &{9.39}\ranktwo  &{6.96}\ranktwo
&11.39 &25.96 &170\\ \hline

\textbf{S²M²-XL} (+2Mtr) &\textbf{2.30}\rankone  &\textbf{7.95}\rankone  &\textbf{9.22}\rankone  &\textbf{5.15}\rankone  &\textbf{3.29}\rankone  &\textbf{2.18}\rankone
 &\textbf{6.2}\rankone  &\textbf{20.58}\rankone  &\textbf{16.51}\rankone  &\textbf{11.60}\rankone  &\textbf{8.56}\rankone  &\textbf{6.36}\rankone
&14.84 &56.94 &381\\ \hline\hline
\end{tabular}}
}
\caption{
Quantitative results on our synthetic benchmark, evaluated over the full disparity range. All models process the input images at their original resolution of 2448~$\times$~2048 pixels. The last three columns report peak memory usage, total computational cost (in tera floating-point operations, TFLOPs), and model size (in millions of parameters). The three best-performing models for each metric are marked with \textcolor{blue}{blue} superscripts.
}
\label{tab:method_comparison_full}
\end{table*}

\begin{table*}[!htp]
\centering
\scriptsize{{
\begin{tabular}{l|l l l l l l|l l l l l l}
\hline\hline
\multirow{2}{*}{\textbf{Model}} & \multicolumn{6}{c|}{\textbf{Non-Occlusion}} & \multicolumn{6}{c}{\textbf{All Pixels}} \\ 
\cline{2-13} 
& \textbf{EPE} & \textbf{RMS} & \textbf{Bad-0.5} & \textbf{Bad-1.0} & \textbf{Bad-2.0} & \textbf{Bad-4.0} 
& \textbf{EPE} & \textbf{RMS} & \textbf{Bad-0.5} & \textbf{Bad-1.0} & \textbf{Bad-2.0} & \textbf{Bad-4.0} \\ 
\hline\hline
{RAFTStereo} \cite{lipson2021raft} & 0.18 & 0.36 & 7.04  & 2.44   & 0.44   & 0.15
                                          & 0.19 & 0.42 & 7.33  & 2.60   & 0.56   & 0.22 \\ \hline
{CREStereo} \cite{li2022practical} & 0.13 & 0.28 & 3.58  & 0.98   & 0.22   & 0.10
                                          & 0.14 & 0.31\rankthree & 3.75  & 1.09   & 0.29   & 0.12 \\ \hline
{Unimatch} \cite{xu2023unifying}   & 0.19 & 0.38 & 5.94  & 1.83   & 0.25   & 0.08
                                          & 0.21 & 0.44 & 6.44  & 2.07   & 0.34   & 0.14 \\ \hline
{CroCo} \cite{weinzaepfel2023croco} & 0.14 & 0.30 & 3.27  & 0.99   & 0.39   & 0.13
                                          & 0.15 & 0.35 & 3.51  & 1.14   & 0.50   & 0.18 \\ \hline
{IGEV} \cite{xu2023iterative}    & 0.14 & 0.34 & 3.52  & 1.12   & 0.21   & 0.11
                                          & 0.20 & 0.86 & 3.97  & 1.51   & 0.54   & 0.41 \\ \hline
{Sel-IGEV} \cite{wang2024selective} & 0.12  & 0.29 & 3.06  & 1.23   & 0.22   & 0.05
                                          & 0.15 & 0.57 & 3.46  & 1.56   & 0.51   & 0.28 \\ \hline                                   
{CAS}  \cite{min2024confidence}    & 0.14 & 0.30 & 3.96  & 0.87   & 0.32   & 0.07
                                          & 0.15 & 0.32 & 4.11  & 1.09   & 0.21\rankthree   & 0.10\rankthree \\ \hline
{Mocha V2} \cite{chen2024motifv2}  & 0.13\rankthree & 0.36 & 3.20  & 1.41   & 0.33   & 0.09
                                          & 0.18 & 0.74 & 3.68  & 1.83   & 0.71   & 0.41 \\ \hline   
\textbf{S²M²-L} (+1Mtr) & 0.13\rankthree  & 0.24\rankthree & 2.06\rankthree  & 0.46\rankthree   & 0.13\rankthree   & 0.05\ranktwo
                               & {0.13}\ranktwo & 0.27\ranktwo & 2.24\rankthree  & 0.53\rankthree   & 0.17\ranktwo   & 0.08\ranktwo \\ \hline\hline

{FoundationStereo \cite{wen2025foundationstereo} (+2Mtr)}     & \textbf{0.09}\rankone & \textbf{0.20}\rankone & 1.26\ranktwo  & 0.26\ranktwo   & 0.08\ranktwo   & 0.05\ranktwo
                                                              & 0.13\ranktwo & 0.61 & 1.56\ranktwo  & 0.48\ranktwo   & 0.26   & 0.21 \\ \hline

\textbf{S²M²-XL} (+2Mtr)  & 0.10\ranktwo & \textbf{0.20}\rankone & \textbf{0.93}\rankone & \textbf{0.22}\rankone   & \textbf{0.06}\rankone   & \textbf{0.03}\rankone
                                    & \textbf{0.10}\rankone & \textbf{0.22}\rankone & \textbf{1.03}\rankone  & \textbf{0.26}\rankone   & \textbf{0.08}\rankone   & \textbf{0.04}\rankone  \\ \hline\hline                      
\end{tabular}}
}
\caption{Quantitative results on the ETH3D stereo benchmark~\cite{schops2017multi}. The three best-performing models for each metric are marked with \textcolor{blue}{blue} superscripts.}
\label{table:ETH}
\end{table*}

\begin{table*}[!hpt]
\centering
\scriptsize{{
\begin{tabular}{l|l l l l l l|l l l l l l}
\hline\hline
\multirow{2}{*}{\textbf{Model}} & \multicolumn{6}{c|}{\textbf{Non-Occlusion}} & \multicolumn{6}{c}{\textbf{All Pixels}} \\ 
\cline{2-13} 
& \textbf{EPE} & \textbf{RMS} & \textbf{Bad-0.5} & \textbf{Bad-1.0} & \textbf{Bad-2.0} & \textbf{Bad-4.0} 
& \textbf{EPE} & \textbf{RMS} & \textbf{Bad-0.5} & \textbf{Bad-1.0} & \textbf{Bad-2.0} & \textbf{Bad-4.0} \\ 
\hline\hline
{RAFTStereo} \cite{lipson2021raft}                & 1.27  & 8.41  & 27.7  & 9.37   & 4.74   & 2.75
                                                  & 2.71  & 12.6  & 33.0  & 15.1   & 9.37   & 6.42 \\ \hline
{CREStereo} \cite{li2022practical}                & 1.15  & 7.70  & 33.3  & 8.25   & 3.71   & 2.04
                                                  & 2.10  & 10.5  & 31.3  & 14.0   & 8.13   & 5.05 \\ \hline
{Unimatch} \cite{xu2023unifying}                  & 1.31  & 6.45\rankthree  & 51.5  & 23.6   & 7.14   & 2.96
                                                  & 1.89  & 8.03\ranktwo  & 54.8  & 28.4   & 11.7   & 6.07 \\ \hline
{CroCo} \cite{weinzaepfel2023croco}               & 1.76  & 8.91  & 40.6  & 16.9   & 7.29   & 4.18
                                                  & 2.36  & 10.6  & 44.4  & 21.6   & 11.1   & 6.75 \\ \hline
{IGEV} \cite{xu2023iterative}                     & 2.89  & 12.8  & 32.4  & 9.41   & 4.83   & 3.33  
                                                  & 3.64  & 15.1  & 31.3  & 13.8   & 8.16   & 5.79 \\ \hline
{Sel-IGEV} \cite{wang2024selective}               & 0.91  & 7.26  & 24.6\rankthree  & 6.53   & 2.51   & 1.36
                                                  & 1.54  & 9.26  & 36.6  & 11.4   & 6.04   & 3.74 \\ \hline
{CAS}  \cite{min2024confidence}                   & 1.01  & 6.74  & 31.3  & 8.71   & 3.33   & 1.77
                                                  & 1.52  & 8.54  & 35.3  & 13.0   & 6.29   & 3.75 \\ \hline
{Mocha V2} \cite{chen2024motifv2}                 & 1.10  & 7.69  & 24.8  & 6.81   & 3.51   & 2.34
                                                  & 1.83  & 9.91  & 29.4\rankthree  & 11.4   & 6.74   & 4.54 \\ \hline
\textbf{S²M²-L} (+1Mtr)                 & 0.80\rankthree  & 6.36\ranktwo  & 25.5  & 6.31\rankthree   & 2.08\rankthree   & 0.97\ranktwo
                                                  & 1.30\rankthree  & 8.31\rankthree  & 29.5  & 10.2\rankthree   & 4.76\rankthree	  & 2.85\rankthree \\ \hline\hline

{FoundationStereo \cite{wen2025foundationstereo} (+2Mtr)}     & 0.78\ranktwo  & 6.48  & 22.5\ranktwo  & 4.39\ranktwo   & 1.84\ranktwo   & 1.04\rankthree
                                                              & 1.24\ranktwo  & 8.39  & 26.4\ranktwo  & 8.14\ranktwo   & 4.26\ranktwo   & 2.72\ranktwo \\ \hline
\textbf{S²M²-XL} (+2Mtr)                            & \textbf{0.69}\rankone  & \textbf{6.01}\rankone  & \textbf{22.1}\rankone  & \textbf{3.57}\rankone   & \textbf{1.15}\rankone   & \textbf{0.54}\rankone
                                                              & \textbf{1.00}\rankone  & \textbf{7.39}\rankone  & \textbf{25.3}\rankone  & \textbf{6.60}\rankone   & \textbf{2.94}\rankone	  & \textbf{1.63}\rankone
\\ \hline\hline

\end{tabular}}
}
\caption{Quantitative results on the Middlebury v3 Benchmark~\cite{scharstein2014high}.  The three best-performing models for each metric are marked with \textcolor{blue}{blue} superscripts.}
\label{table:middlebury}
\end{table*}

\begin{table*}[!htp]
\centering
\scriptsize{
\begin{tabular}{l c |l l l |l l l}
\hline\hline
\multirow{2}{*}{\textbf{Model}} & \multirow{2}{*}{\textbf{Fine-tune}} & \multicolumn{3}{c|}{\textbf{Non-Occlusion}} & \multicolumn{3}{c}{\textbf{All Pixels}} \\ 
\cline{3-8} 
& & \textbf{D1-bg} & \textbf{D1-fg} & \textbf{D1-all} 
& \textbf{D1-bg} & \textbf{D1-fg} & \textbf{D1-all} \\ 
\hline\hline
{Sel-IGEV}~\cite{wang2024selective} & Yes & 1.22	&2.55	&1.44 &  1.33 &	2.61 &	1.55 \\ \hline
{Sel-IGEV}~\cite{wang2024selective} & No & 2.91 & 12.9	& 4.57 & 3.06 &	13.4 &	4.79  \\ \hline
{FoundationStereo}~\cite{wen2025foundationstereo} & No & 2.48 & 6.13 &	3.08 & 2.59 &	6.25 & 3.20  \\ \hline
\textbf{S²M²-XL}   & No & 2.51 &	6.22&	3.12& 2.61&	6.31&	3.23  \\ \hline
\end{tabular}
}
\caption{Quantitative results on the KITTI 2015 Benchmark~\cite{menze2015object}. The results highlight the dramatic performance gap between models with and without dataset-specific fine-tuning, suggesting that top leaderboard scores are heavily influenced by adaptation to dataset biases rather than true generalization. }
\label{table:KITTI_benchmark}
\end{table*}

\section{Experiments}

\subsection{Implementation Details}
All models were trained using the AdamW optimizer, an initial learning rate of $1\times10^{-5}$, a one-cycle schedule, and three local refinement iterations. For the ablation studies, we used a 0.5M image subset from several synthetic datasets~\cite{mayer2016large,wang2020tartanair,li2022practical}. Our final models for the benchmark evaluation were trained on an expanded 2M image dataset (denoted `+2Mtr`) that includes a diverse mix of data~\cite{mayer2016large, wang2020tartanair, li2022practical, wang2021irs, tremblay2018falling, gaidon2016virtual, yang2019drivingstereo, tosi2021smd, min2024confidence, wen2025foundationstereo}, and subsequently underwent multi-resolution fine-tuning. All experiments were conducted on an Nvidia H100 GPU, and models were evaluated at their original resolutions. Further details on data processing and augmentations are provided in the supplementary material.

All evaluations are conducted using standard stereo matching metrics to assess disparity accuracy, including End-Point Error (EPE), Root Mean Square Error (RMS), and Bad-p thresholds (Bad-0.5, Bad-1.0, Bad-2.0, Bad-4.0). Each metric is reported for both non-occluded regions and all pixels, where lower values indicate better performance. For confidence and occlusion estimation, we adopt an Average Precision (AP) evaluation approach, where values closer to 1 indicate better performance.

\subsection{Ablation Study}
We conducted an ablation study on the Sceneflow dataset~\cite{mayer2016large} to validate our architectural design choices, with results summarized in Table~\ref{tab:ablation} and Table~\ref{tab:refinement}. First, we confirmed the scalability of our architecture: increasing the channel dimension (Ch) and number of Transformer blocks (Ntr) consistently improves accuracy. Next, we validated the effectiveness of our key components, showing that the Gated Fusion mechanism enhances feature aggregation and the Probabilistic Mode Concentration (PMC) loss boosts accuracy without added computational cost. Finally, we analyzed the refinement process and found that S²M² requires far fewer iterations than prior methods. Unlike approaches reliant on many updates, our model's strong initial disparity estimate allows it to achieve near-optimal performance with just three refinement steps, demonstrating the effectiveness of our global matching foundation.

\subsection{Benchmark Evaluation}

We evaluate S²M² on diverse benchmarks to verify its effectiveness across realistic and challenging stereo conditions. To thoroughly assess its capabilities and study scaling effects, we introduce several variants of S²M². These differ in model size (S: Ch=128, Ntr=1; L: Ch=256, Ntr=3; XL: Ch=384, Ntr=3) and are trained with datasets of varying scales, including conventional-sized datasets (denoted `+1Mtr`) and expanded datasets of approximately 2 million stereo pairs (denoted `+2Mtr`), including the FoundationStereo dataset\cite{wen2025foundationstereo}. This multi-faceted approach allows for a comprehensive analysis of performance scaling with respect to both model capacity and training data volume. Specifically, we conduct comparisons on our synthetic dataset, as well as on established real-world datasets such as ETH3D~\cite{schops2017multi}, Middlebury v3~\cite{scharstein2014high}, and the KITTI 2015 benchmark~\cite{menze2015object}.

\subsubsection*{Our Synthetic Benchmark}

Our synthetic benchmark is specifically designed to assess performance in challenging high-resolution, large-disparity scenarios that push modern methods to their limits. This setting proves difficult for many existing approaches, as shown in Table~\ref{tab:method_comparison_full}. For instance, methods relying on explicit cost volumes (e.g., IGEV~\cite{xu2023iterative}, FoundationStereo~\cite{wen2025foundationstereo}, Mocha~\cite{chen2024mocha}) are constrained by a maximum disparity, which we capped at 960 pixels due to their memory requirements. While this covers over 99$\%$ of the disparities in our test set, it highlights a fundamental architectural limitation. Unsurprisingly, these models, as well as lightweight approaches~\cite{bangunharcana2021correlate,xu2021bilateral,xu2023accurate,feng2024mc,guan2024neural}, struggle to generalize in this demanding environment.

\begin{figure}[!htbp]
    \centering
    \includegraphics[width=0.95\linewidth]{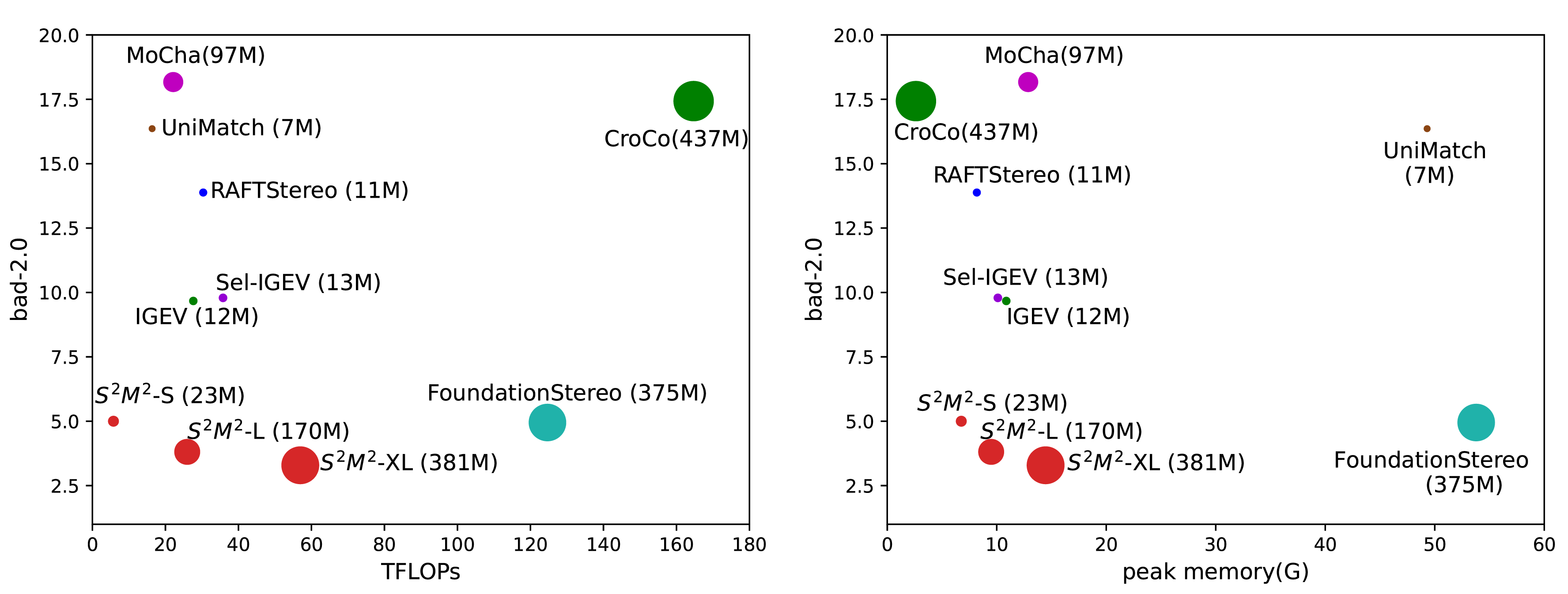}
    \vspace*{-1mm}
    \captionof{figure}{Scalability Analysis}
    \label{fig:scalability}
\end{figure}

In stark contrast, our S²M² family demonstrates exceptional robustness and a superior accuracy-efficiency trade-off. Our largest model, S²M²-XL (+2Mtr), establishes a new state-of-the-art across all metrics. While FoundationStereo~\cite{wen2025foundationstereo} achieves the best performance among prior methods, it is computationally prohibitive. This trade-off is visualized in the scatter plot of Figure~\ref{fig:scalability}, which plots accuracy against model size. The figure clearly shows FoundationStereo~\cite{wen2025foundationstereo} occupying a high-accuracy but high-cost region, whereas our S²M² family forms a compelling Pareto front, offering significantly better performance at every level of computational budget and validating the scalability of our architecture.

\subsubsection*{ETH3D Benchmark \cite{schops2017multi}}
The ETH3D benchmark~\cite{schops2017multi} is characterized by real-world scenes with predominantly small disparity values, making it an excellent testbed for assessing precision and robustness in a different regime from our large-disparity synthetic data. On this benchmark, S²M² once again demonstrates its state-of-the-art capabilities, as shown in Table~\ref{table:ETH}.

Our S²M²-XL model surpasses prior methods like FoundationStereo~\cite{wen2025foundationstereo} across the majority of metrics. Achieving top performance on both our large-disparity synthetic dataset and the small-disparity ETH3D benchmark validates the exceptional versatility of our global matching architecture across a wide spectrum of disparity ranges. The model's ability to maintain high accuracy in ETH3D's varied conditions, including challenging low-texture surfaces and noisy scenes, further highlights its practical robustness.

\subsubsection*{The Middlebury v3 Benchmark \cite{scharstein2014high}}
The Middlebury v3 benchmark~\cite{scharstein2014high}, with its high-resolution images and precise ground-truth, serves as the ultimate test for a model's ability to preserve fine-grained details. On this demanding benchmark, S²M² again demonstrates the superiority of our approach, establishing a new state-of-the-art as detailed in Table~\ref{table:middlebury}.

Our S²M²-XL model not only outperforms strong baselines like FoundationStereo~\cite{wen2025foundationstereo}, but does so with particularly large margins on the most stringent metrics (Bad-1.0 and Bad-2.0). This numerical superiority translates to tangible qualitative improvements. For instance, as seen in Figures~\ref{fig:thumbnail}, our model faithfully reconstructs highly intricate structures like bicycle spokes—a common failure case for many other methods. This result indicates that our global matching framework is effective at preserving fine details by establishing accurate initial correspondences for subsequent refinement.

\subsubsection*{Critical Re-evaluation of the KITTI Benchmark~\cite{Geiger2012CVPR, menze2015object}}
While the KITTI benchmark~\cite{menze2015object} is a standard for stereo evaluation, we argue its leaderboard scores are an unreliable indicator of true generalization due to inherent noise in its LiDAR-based ground truth. Our analysis in Table~\ref{table:KITTI_benchmark} highlights this discrepancy. Powerful generalist models like FoundationStereo~\cite{wen2025foundationstereo} and our S²M²-XL, despite their reliable visual quality and top performance on other benchmarks, exhibit error rates on KITTI that are more than double those of the top fine-tuned models on the leaderboard. As seen in Table~\ref{table:KITTI_benchmark}, this vast performance gap suggests that top KITTI scores are less a measure of a model's inherent capability and more a reflection of its ability to adapt to dataset-specific biases and noises.

Our supplementary material provides extensive evidence supporting this claim. We demonstrate that while fine-tuning improves error metrics, it simultaneously degrades photometric similarity measured via left-right warping. Furthermore, we provide 3D visualizations that show fine-tuned models adapting to noise patterns, and present manual inspection results identifying consistent ground-truth biases of up to 2 pixels. Therefore, we recommend that KITTI leaderboard scores be interpreted with caution, as they may disproportionately reward models that fit to flawed annotations rather than those that achieve genuine real-world accuracy.

\section{Conclusion}

This paper presents S²M², a multi-resolution Transformer framework that overcomes the long-standing challenges of global matching, positioning it as a practical, state-of-the-art methodology. The proposed framework is highly scalable, robustly handling challenging high resolutions and wide disparity ranges, while its performance improves consistently with model size. It produces reliable depth estimates by jointly modeling disparity, confidence, and occlusion. Our extensive experiments not only establish a new state-of-the-art on real-world benchmarks but also demonstrate the framework's robustness in demanding scenarios, thereby opening new avenues for developing highly accurate and generalizable stereo systems.


\newpage
{\small
\bibliographystyle{ieeenat_fullname}
\bibliography{refbib}
}

\end{document}